\documentclass{article}

\usepackage{arxiv}

\usepackage[utf8]{inputenc} 
\usepackage[T1]{fontenc}    
\usepackage{hyperref}       
\usepackage{url}            
\usepackage{booktabs} 
\usepackage{amssymb}
\usepackage{amsmath}      
\usepackage{amsfonts}       
\usepackage{nicefrac}       
\usepackage{microtype}      
\usepackage{cleveref}       
\usepackage{lipsum}         
\usepackage{graphicx}
\usepackage{natbib}
\usepackage{doi}
\usepackage{hyperref}
\usepackage{xcolor}

\title{FixationFormer: Direct Utilization of Expert Gaze Trajectories for Chest X-Ray Classification}

\date{}

\author{ \href{https://orcid.org/0000-0002-1233-7730}{\includegraphics[scale=0.06]{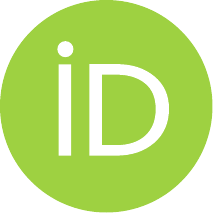}\hspace{1mm}Daniel~Beckmann}\\
	Institute for Geoinformatics \\
	Department of Computer Science\\
	University of Münster\\
	Münster, Germany \\
	\texttt{daniel.beckmann@uni-muenster.de} \\
	\And
	\href{https://orcid.org/0000-0001-5691-4029}{\includegraphics[scale=0.06]{orcid.pdf}\hspace{1mm}Benjamin~Risse\thanks{Corresponding Author}} \\
	Institute for Geoinformatics \\
	Department of Computer Science\\
	University of Münster\\
	Münster, Germany \\
	\texttt{b.risse@uni-muenster.de} \\
}

\renewcommand{\headeright}{}
\renewcommand{\undertitle}{}
\renewcommand{\shorttitle}{FixationFormer}

\hypersetup{
pdftitle={FixationFormer},
pdfsubject={cs.CV},
pdfauthor={Daniel~Beckmann, Benjamin~Risse},
pdfkeywords={Eye Tracking, Chest X-Ray, Machine Learning},
}

\begin{document}
\maketitle

\begin{abstract} 
Expert eye movements provide a rich, passive source of domain knowledge in radiology, offering a powerful cue for integrating diagnostic reasoning into computer-aided analysis.
However, direct integration into CNN-based systems, which historically have dominated the medical image analysis domain, is challenging: gaze recordings are sequential, temporally dense yet spatially sparse, noisy, and variable across experts.
As a consequence, most existing image-based models utilize reduced representations such as heatmaps. 
In contrast, gaze naturally aligns with transformer architectures, as both are sequential in nature and rely on attention to highlight relevant input regions.
In this work, we introduce FixationFormer, a transformer-based architecture that represents expert gaze trajectories as sequences of tokens, thereby preserving their temporal and spatial structure.
By modeling gaze sequences jointly with image features, our approach addresses sparsity and variability in gaze data while enabling a more direct and fine-grained integration of expert diagnostic cues through explicit cross-attention between the image and gaze token sequences.
We evaluate our method on three publicly available benchmark chest X-ray datasets and demonstrate that it achieves state-of-the-art classification performance, highlighting the value of representing gaze as a sequence in transformer-based medical image analysis.
Code is available at \url{https://zivgitlab.uni-muenster.de/cvmls/fixation_former}.

\keywords{Eye-Tracking  \and Transformer \and Chest X-Ray Classification}
\end{abstract}
    
\section{Introduction}
\label{sec:intro}

Medical image analysis has traditionally relied heavily on Convolutional Neural Networks (CNNs), which have achieved impressive results in various applications, including disease diagnosis and image segmentation~\citep{zhou2018unetnestedunetarchitecture, isensee2021nnunet}.
However, medical image datasets are often smaller and more complex than those used for natural images, posing significant challenges for model development and training.
One such challenging domain is the analysis of chest X-rays images.
For this modality used in clinical routine, the chest region of the patient is irradiated with X-Rays, producing a 2D projection of the entire area showing organs, bone and tissue structure.
The projection leads to overlapping organs and regions of interest (ROIs) on the image, making accurate analysis difficult.
To address these challenges, researchers have started exploring enriching image-based models with auxiliary information, with gaze information emerging as a promising example.
Several studies have shown that incorporating gaze from domain experts, in this case trained radiologists, can improve performance in medical image analysis tasks, such as detecting abnormalities and diagnosing diseases~\citep{karargyris_creation_2021, zhu_gaze-guided_2022, saab_observational_2021, wang_follow_2022, ma_eye-gaze-guided_2022, wang_gazegnn_2024}.
Gaze can provide valuable insights into how medical professionals analyze images, highlighting areas of interest and helping models focus on relevant features.

Given the dominance of CNN models in medical image analysis, the gaze information has typically been incorporated by using image-based representations, such as heatmaps, which reduce sequential gaze patterns to a 2D attention map~\citep{karargyris_creation_2021, zhu_gaze-guided_2022, saab_observational_2021, wang_follow_2022}.
However, 2D heatmaps fail to preserve the temporal dynamics of gaze patterns, which may convey important contextual information, while their computation can also be expensive~\citep{wang_gazegnn_2024}.

More recently, transformer-based models have demonstrated superior performance across a wide range of image analysis tasks, most notably with the introduction of the Vision Transformer (ViT), which treats images as sequences of 16×16 patches, or “visual words”~\citep{dosovitskiy2021imageworth16x16words}. 
Beyond their success in computer vision, transformers have a natural conceptual connection to gaze data, as both rely on attention mechanisms to focus selectively on relevant input regions. 
Moreover, transformers were originally designed for sequential data, making them particularly well suited to model gaze trajectories, which are inherently sequential in nature. 

Some studies have begun to explore this direction in medical imaging, for example by using gaze as a prompt for segmentation models such as SAM~\citep{wang2023gazesam, beckmann2023sam} and MedSAM~\citep{khaertdinova_gaze-assisted_2024, shmykova_zero-shot_2025}. Both employ a ViT backbone for image encoding and have shown promising results in specific applications, such as weakly supervised image annotation.
For both models, gaze serves as a proxy for mouse-based prompting, utilizing only the spatial information contained in one or a few gaze points.

This paper proposes a novel approach called \emph{FixationFormer} for integrating gaze into medical image analysis, representing gaze as a sequence of tokens that can be seamlessly integrated into a transformer-based architecture.
In particular, we transform raw gaze trajectories into tokens in two steps.
First, we calculate the fixation sequence for each gaze trajectory.
Then, we derive temporal and spatial encodings from each fixation, fusing them to obtain one token per fixation.
To fuse these with image features extracted by a standard ViT pretrained on the large MIMIC-CXR dataset~\citep{johnson_mimic-cxr_2019}, we implement two complementary attention mechanisms.
The first, Image-to-Gaze Cross-Attention, updates only the image features by attending to the gaze tokens, ensuring that visual representations are enriched with expert viewing patterns. 
The second, Two-Way Attention, extends this design by additionally updating the gaze tokens through mirrored attention, enabling a deeper bidirectional fusion of gaze and image features.

We apply these approaches to chest X-ray classification and evaluate our model based on three benchmark datasets.
Our results demonstrate that this new method can achieve competitive performance, matching or improving upon state-of-the-art results, highlighting the potential of leveraging gaze as a sequence in medical image analysis.
\section{Related  Work}
\label{sec:related_work}

Converting a gaze trajectory into a static heatmap image allows for convenient integration of gaze information into image-based models.
Hence, various previous works have studied this gaze representation for different image domains and corresponding tasks.
Several methods interpret gaze-derived heatmaps as a proxy for human attention and try to align the models' attention with the human one.
Specifically, Koorathota et al.~\citep{koorathota_gaze-informed_2025} improve the image-based prediction of driving decisions by aligning the attention mechanism of a Vision Transformer (ViT)~\citep{dosovitskiy2021imageworth16x16words} with the human attention by incorporating an additional attention intersection loss.
Contrary to this use of gaze as a additional regularizer during training only, Hu et al.~\citep{hu_gazevit_2025} integrate gaze-based attention as an additional input into multi-view matching for street and aerial images.
They modify the attention of a pretrained ViT during a second training stage with a learned hybrid attention module which combines both human and ViT attention.

In the medical domain, Zhu et al.~\citep{zhu_gaze-guided_2022} and Wang et al.~\citep{wang_follow_2022} utilize gaze-based heatmaps for Chest X-Ray classification and Osteoathritis ranking, respectively.
As both methods deploy CNN-based models, they rely on CAM~\citep{zhou2015learningdeepfeaturesdiscriminative} to infer the models' attention. 
During training, the extracted attention map is aligned with the human attention on the pixel level using the MSE loss.

Gaze has also been shown to be beneficial for self-supervised pretraining on images or in multi-modal settings, especially in the medical domain where datasets are usually of substantially smaller size.
Ma et al.~\citep{ma_eye-gaze_2024} introduce Eye-gaze Guided Multi-modal Alignment (EGMA), a CLIP-based~\citep{radford2021learningtransferablevisualmodels} multi-modal pretraining method for Chest X-Ray images and their corresponding medical reports.
Unlike CLIP, which matches multi-modal features on an instance level, EGMA allows for an additional alignment of image regions and single sentences.
To achieve this, the authors utilize the recorded gaze trajectories to map sentences to their corresponding image regions.

Wang et al.~\citep{wang_improving_2025} observed similar gaze patterns on different images when both images share a common semantic feature. 
To improve upon classical contrastive learning, they introduce a trajectory-based gaze similarity measure to automatically identify semantically similar images via their matching gaze patterns. 
This allows them to introduce additional semantic-derived positive pairs within the contrastive learning framework, improving its performance on several medical image datasets.

In contrast to temporally aggregated heatmap representation of gaze, single gaze points or short trajectories have been shown to be useful for prompt-based image segmentation models. 
Following the release of the object-agnostic segmentation framework SAM~\citep{kirillov2023segment} which utilizes clicks or bounding boxes as prompts, Beckmann et al.~\citep{beckmann2023sam} and Wang et al.~\citep{wang2023gazesam} demonstrated that these mouse-based prompting mechanism can effectively be replaced by gaze-based queries.
In the context of medical images, Khaertdinova et al.~\citep{khaertdinova_gaze-assisted_2024} and Shmykova et al.~\citep{shmykova_zero-shot_2025} showed that also the MedSAM model~\citep{MedSAM} and its successor~\citep{MedSAM2} can be finetuned for gaze prompts or used for zero-shot gaze-based 3D medical image segmentation, respectively.

In addition to the aforementioned methods, other integration methods have studied for classification tasks in the medical image domain, specifically for Chest X-Ray disease classification.
Along with utilization of gaze for the creation of weak labels, Saab et al.~\citep{saab_observational_2021} showed that the prediction of gaze-derived statistics as an auxiliary task can improve performance for Chest X-Ray classification. 
Ma et al.~\citep{ma_eye-gaze-guided_2022} introduce EG-ViT, a Vision Transformer based model which uses gaze trajectories to penalize irrelevant image regions on the token level to improve performance.

Several works have studied the integration of gaze information into ViT-based architectures.
Pham et al.~\citep{pham2025itpctrl} use expert gaze to train a promptable model which mimics the radiologists' attention pattern by producing attention heatmaps, which are then used to mask unimportant image regions for downstream tasks.
Chen et al.~\citep{chen2026gaze} propose the integration of gaze information directly into ViTs by adding learned point embeddings to the image tokens in each layer.
These point embeddings are derived from a duration-dependent gaze heatmap by a linear layer.
A student-teacher network is proposed by Bhattacharya et al.~\citep{bhattacharya2022radiotransformer} where the teacher is pretrained to extract visual attention patterns from radiologists' gaze and guides the attention of a student network via a Visual Attention Loss.

Lastly, Wang et al.~\citep{wang_gazegnn_2024} introduced a graph-based classification model called GazeGNN, which fuses gaze- and image-derived features into a graph representation, achieving state-of-the-art classification accuracy on one of the widely adopted Gaze Chest X-Ray datasets~\citep{karargyris_creation_2021}.

To this end, the direct integration of gaze trajectories as a sequence into a Transformer-based model remains largely unexplored.

\begin{figure*}[t!]
	\centering
	\includegraphics[width=\textwidth]{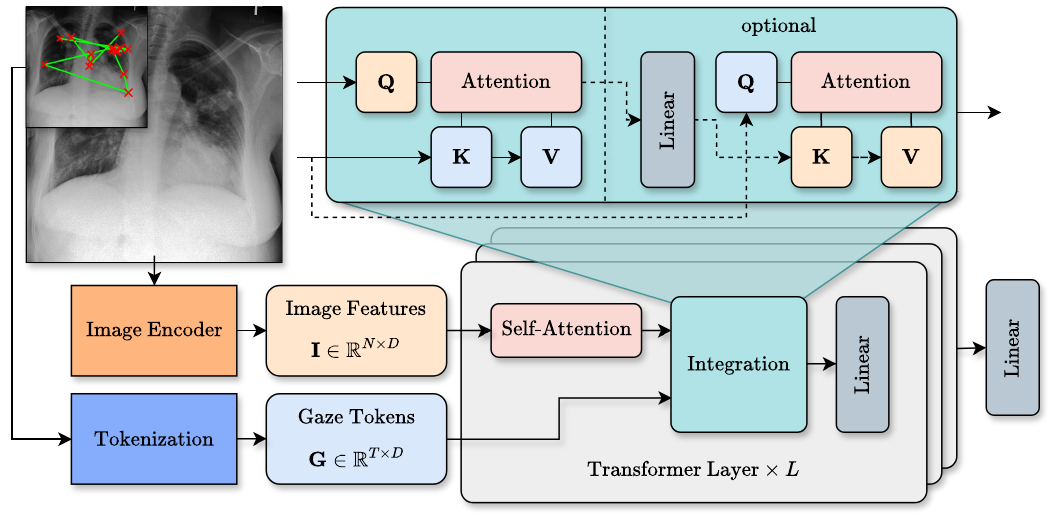}
	\caption{\textbf{FixationFormer overview}: Image and gaze are encoded into separate token sequences. 
		To infuse the image features with gaze information, we use cross-attention in one or optionally both directions throughout a stack of Transformer layers. 
		Finally, the [\texttt{CLS}] token from the image encoder is passed to the prediction head.}
	\label{fig:main_fig}
\end{figure*}
\section{Methods}

\label{sec:methods}
In this section we describe our proposed FixationFormer model for the direct utilization of gaze trajectories. 
Figure~\ref{fig:main_fig} illustrates the proposed architecture comprised of an image backbone, a tokenisation strategy for gaze trajectories as well as the Gaze Integration module.
The following sections describe these building blocks used in more detail.

\subsection{Image Encoder}
\label{ssec:image_encoder}
Our method uses a standard Vision Transformer~\citep{dosovitskiy2021imageworth16x16words} as the image backbone.
In order to compensate for the drop in performance of ViTs when trained on comparatively small datasets~\citep{zhu2023understandingvittrainsbadly, lee2021visiontransformersmallsizedatasets}, we pretrain our image encoder using the Multi-Granularity Cross-modal Alignment framework (MGCA) introduced by Wang et al.~\citep{wang_multi-granularity_2022}.
Specifically, we train a multi-modal model on the large MIMIC-CXR~\citep{johnson_mimic-cxr_2019} Chest X-Ray dataset and use the ViT-based encoder as the image feature backbone.
To avoid overlap between the data used in pretraining and for our experiments, we filter the MIMIC-CXR dataset accordingly before pre-training.

\begin{figure}[h!]
\centering
      \includegraphics[width=0.8\linewidth]{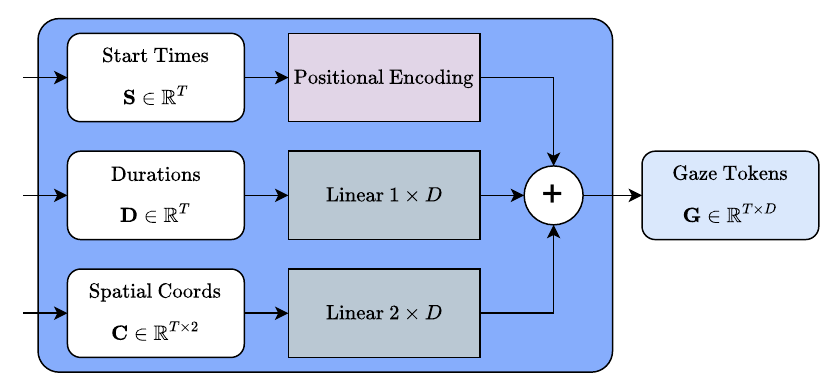}
\caption{\textbf{Gaze Representation}: The fixation trajectory (containing the start time, duration and spatial location for each fixation) is transformed into a token sequence. 
Spatial location and duration are projected using a learned linear layer, while the start times are used to encode the relative positional information using the positional embeddings proposed by~\citep{vaswani2017attentionneed}.} 
\label{fig:gaze_representation}
\end{figure}

\subsection{Gaze Representation}
\label{ssec:gaze_representation}

Raw gaze trajectories, typically recorded at $60$Hz or higher, naturally contain a lot of micro-movements called saccades.
To filter and condense this raw and noisy signal, it is common practice to transform raw gaze into a sequence of fixation points.
Fixations correspond to points of interest where the eyes remain focused on a region for a sustained period, typically on the order of seconds.
This representation reduces noise while preserving most of the spatial and temporal information contained in the gaze, while also significantly shortening trajectory length and thereby reducing computational cost.
We therefore represent gaze trajectories by their constituent fixations, in line with common practice in the field.
For all datasets used in this work, fixations were already calculated and provided alongside the raw gaze trajectories.

A fixation is fully described by its spatial location, its start time within the full trajectory and its duration.
To transform a given sequence of fixations into a sequence of tokens, we calculate a token for each fixation separately as depicted in Figure~\ref{fig:gaze_representation}.
Given a fixation trajectory $\mathbf{F} \in \mathbb{R}^{T\times 4}$ of length $T$, we first encode the durations  $\mathbf{D} \in \mathbb{R}^{T \times 1}$ and spatial coordinates  $\mathbf{C} \in \mathbb{R}^{T\times 2}$ into the $D$-dimensional token space using using learnable linear transformations $\mathbf{L}_\mathbf{D} \in \mathbb{R}^{1\times D}$ and $\mathbf{L}_\mathbf{C} \in \mathbb{R}^{2\times D}$.
The start times  $\mathbf{S} \in \mathbb{R}^{T \times 1}$ (in seconds) are used to encode the temporal positions of each fixation.
We use the positional encoding from the original Transformer architecture~\citep{vaswani2017attentionneed}.
The final gaze token sequence $\mathbf{G} \in \mathbb{R}^{T\times D}$ is obtained by adding the three representations:
\begin{equation*}
    \mathbf{G} = \mathrm{PositionalEncoding}(\mathbf{S}) + \mathbf{L}_\mathbf{D}\mathbf{D} + \mathbf{L}_\mathbf{C}\mathbf{C}
\end{equation*}

\subsection{Integration with Image Features}
\label{ssec:integration_methods}

\begin{figure*}[ht!]
\centering
      \includegraphics[width=\textwidth]{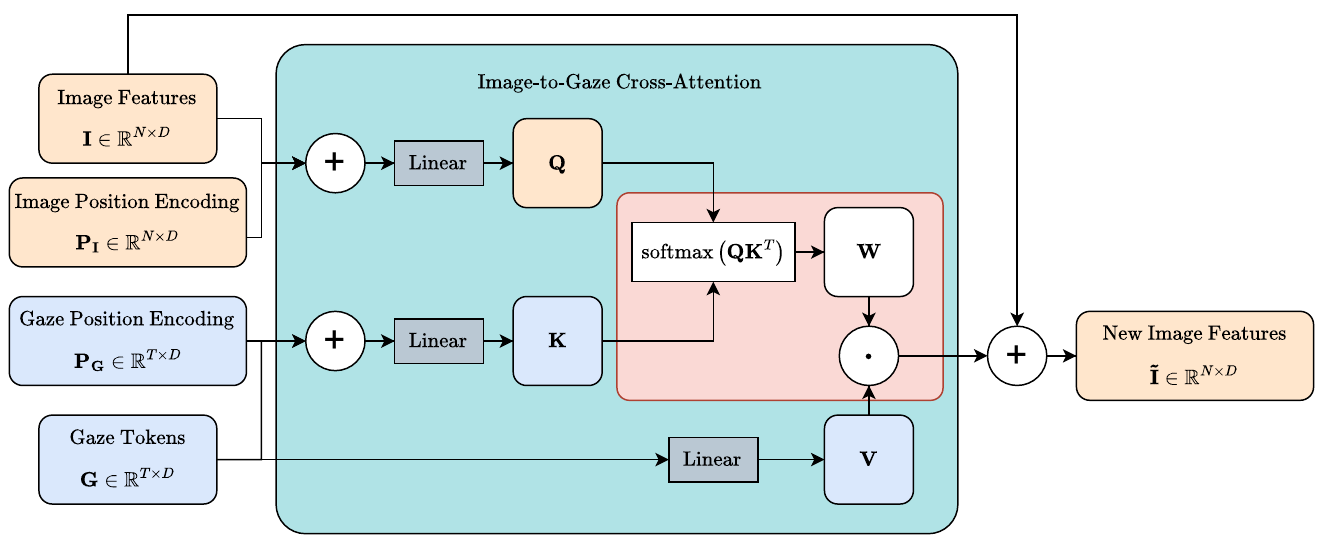}
\caption{\textbf{Image-to-Gaze Cross-Attention}: Images and gaze features are fused by cross-attention. 
Output features of the attention mechanism are based on gaze features, but the skip-connection allows for conservation of image information in the output tokens. 
No masking is applied to allow access to all gaze tokens for each image patch.} 
\label{fig:gaze_integration}
\end{figure*}

For our Gaze Integration module, we use cross-attention to fuse the information from the image and gaze features.
Specifically, the Gaze Integration Module consists of a stack of decoder-style Transformer layers.
As depicted in Figure~\ref{fig:main_fig}, in each layer, image features attend to themselves using self-attention before being infused with information from the gaze token sequence.
To realize this fusion, we investigate two alternative attention mechanisms.

\paragraph{Image-to-Gaze Cross-Attention}
In this version (short Cross-Attention), only the image features are updated throughout the Gaze Integration module, as illustrated in Figure~\ref{fig:gaze_integration}.
Following~\citep{kirillov2023segment} and in order to emphasize the spatial correlation of image regions and gaze, we add an additional spatial positional encoding to both image and gaze features before cross-attention.
This explicit re-introduction of spatial encoding in every layer ensures that the spatial relation between image and gaze tokens remains accessible to the model throughout.

\paragraph{Two-Way Attention}
The second integration method extends the Image-to-Gaze Cross-Attention with a mirrored Gaze-to-Image Cross-Attention, as illustrated in Figure~\ref{fig:main_fig} (top right), forming a Two-Way Attention.
This design closely follows the design of the mask decoder of SAM~\citep{kirillov2023segment}, which fuses information from the image embeddings and prompt tokens.
In contrast to the Image-to-Gaze Cross Attention, this two-way approach allows for the gaze tokens to be updated throughout the Gaze Integration module as well, resulting in a deeper fusion of gaze and image features.

Unlike in the standard decoder layer, we do not apply a subsequent mask within the attention mechanism.
In NLP tasks, this usually is employed to prevent output tokens from attending to "future" tokens.
However, for our application, we explicitly allow each image token to gather information from the whole gaze trajectory, and vice versa.
\section{Experiments}
\label{sec:experiments}

This section describes the datasets, evaluation tasks, and training details for all experiments. 
Implementation details are provided in our publicly available code.

\subsection{Datasets}
\label{ssec:datasets}
We evaluate our method on three publicly available datasets for Chest X-Ray classification, which all contain expert gaze. 
Specifically, we use the CXR-Gaze dataset~\citep{karargyris_creation_2021}, the SIIM-ACR challenge dataset~\citep{siim-acr-pneumothorax-segmentation} with gaze data provided by Saab et al.~\citep{saab_observational_2021} and the Reflacx dataset~\citep{bigolin_lanfredi_reflacx_2022}. 
The CXR-Gaze dataset contains roughly $1,000$ images, split equally into the three categories "CHF" (Congestive Heart Failure), "Pneumonia" and "Normal".
The SIIM-ACR challenge and corresponding dataset, comprised of over $3,000$ images, poses the binary classification problem of predicting the presence or absence of Pneumothorax. 
Classes for this dataset are imbalanced by a factor of $\approx 4$ towards absence of Pneumothorax.
For the Reflacx dataset ( $\approx 2,600$ images), a hierarchy of labels based on conditions and abnormalities present in each image is provided by the authors.
We formulate a non-hierarchical classification problem by using the top-level label with the highest score for each image only.

\subsection{Training Details}
\label{ssec:training}
All models are trained for $50$ epochs using the AdamW optimizer~\citep{loshchilov2019decoupledweightdecayregularization} with an initial learning rate of $2\cdot10^{-4}$, weight decay of $0.01$ and a batch size of $64$.
We decrease the learning rate throughout training using a cosine learning rate scheduler.
All experiments reproducing results from other works use hyperparameters as described by the authors.
For the image encoder we deploy a ViT/B-32 model, and use an image size of $224\times224$ pixels during pretraining and finetuning.
The Gaze Integration module inherits the model size from the image backbone, resulting in the same feature dimension, number of heads and layers.
During training on the classification tasks, we do not update the weights of the image encoder directly, but instead apply LoRA~\citep{hu2021loralowrankadaptationlarge} for task-specific finetuning.

We split all datasets into three subsets for training, validation and testing.
If only a training and test split is available in the original dataset, we exclude a subset of the training set for validation.
Results are reported on the test dataset using validation performance as an early stopping criterion.

We implement our experiments in PyTorch~\citep{paszke2019pytorchimperativestylehighperformance} and use the Nested Tensor functionality to efficiently handle the various lengths of fixation sequences.
Nested tensors allow for batching of tensors with unequal size in one or more dimensions, eliminating the need for padding and the use of padding masks in Transformers.
In our setting, where gaze trajectory lengths can vary substantially, this leads to a significant reduction of memory consumption and increased computational efficiency.

\begin{table*}[ht!]
    \centering
    \resizebox{\textwidth}{!}{
\begin{tabular}{l|c|c|c||c|c|c||c|c|c}
    \hline
    & \multicolumn{3}{c||}{CXR-Gaze} & \multicolumn{3}{c||}{SIIM-ACR} & \multicolumn{3}{c|}{Reflacx} \\
    Architecture & Accuracy & F1 & AUC & Accuracy & F1 & AUC & Accuracy & F1 & AUC \\
    \hline
    \hline
    GG-CAM~\cite{zhu_gaze-guided_2022} (from~\cite{wang_gazegnn_2024})      & \multicolumn{1}{|l|}{77.57} & 0.770 & 0.888 &-&-&-&-&-&-\\
    GazeMTL~\cite{saab_observational_2021} (from~\cite{wang_gazegnn_2024})  & \multicolumn{1}{|l|}{78.50} & 0.779 & 0.887 &-&-&-&-&-&-\\
    U-Net + Gaze~\cite{karargyris_creation_2021}&   -   &  -    & 0.870      & \multicolumn{1}{|l|}{81.10} & 0.803 & 0.689 &-&-&-\\
    EG-ViT~\cite{ma_eye-gaze-guided_2022}       &-&0.807&0.909& \multicolumn{1}{|l|}{\textbf{85.60}} & \textbf{0.849} & 0.741 &-&-&-\\
    GII-ViT~\cite{chen2026gaze}                 & -     & 0.806 & 0.919 &-&-&-&-&-&-\\
     \textcolor{gray}{GazeGNN~\cite{wang_gazegnn_2024}}$¹$ & \multicolumn{1}{|l|}{\textcolor{gray}{83.18}}     & \textcolor{gray}{0.823} & \textcolor{gray}{0.923} &-&-&-&-&-&- \\
         GazeGNN~\cite{wang_gazegnn_2024} (repr.)$^1$   & 71.02 \scriptsize $\pm$ 5.36  & 0.700 & 0.879 & 81.84 \scriptsize $\pm$ 1.76& 0.700 & 0.851 & 64.51 \scriptsize $\pm$ 2.37& 0.453 & 0.757\\
    \hline
    Cross-Attention                             & \textbf{84.11} \scriptsize $\pm$ 1.14  & \textbf{0.833} & 0.944 & 84.96 \scriptsize $\pm$ 1.56  & 0.765 & 0.902 & \textbf{70.06} \scriptsize $\pm$ 0.93 & \textbf{0.561} & \textbf{0.853} \\
    Two-Way Attention                           & 82.80 \scriptsize $\pm$ 1.07  & 0.819 & \textbf{0.952} & \textbf{86.40} \scriptsize $\pm$ 1.56  & 0.797 & \textbf{0.915} & 68.06 \scriptsize $\pm$ 2.32 & 0.510 & 0.842 \\ 
\end{tabular}
}
    \caption{\textbf{Classification Performance:} We report metrics on the official test set of each dataset. 
    Standard deviations are calculated over five randomly initialized runs. 
    For brevity, we exclude standard deviations for F1-Score and AUC. 
    $^1$Note that the differences from the original GazeGNN results can be attributed to a different, but consistent, training protocol used for fairness and comparability. For details see Section~\ref{ssec:sota_comp}.}
    \label{tab:main_results}
\end{table*}

\section{Results}
\label{sec:results}

In this section, we present the classification results and analyze the effectiveness of our proposed Gaze Integration module.
To address the latter, we perform an ablation study showing the impact of the image and gaze encoder in isolation.
Secondly, we investigate the benefits of additional gaze under constrained training settings.
In particular, we replace the task-specific MGCA image backbone with a weaker ImageNet ViT.
Finally, we qualitatively evaluate the correlation of our models' attention with the provided expert gaze trajectories.
\subsection{Classification Performance}
\label{ssec:sota_comp}
Results on the three datasets described in Section~\ref{ssec:datasets} are presented in Table~\ref{tab:main_results}.
Performance is evaluated using the overall accuracy, F1-Score and the Receiver operating characteristic Area under Curve (ROC-AUC or AUC).

\paragraph{CXR-Gaze Dataset}
On the CXR-Gaze dataset, both variants of our FixationFormer perform better than current state-of-the-art.
To ensure full comparability between our method and the previous best method, GazeGNN~\citep{wang_gazegnn_2024}, we evaluate both architectures under both training protocols.
Specifically, Wang et al.~\citep{wang_gazegnn_2024} report best results on the test dataset across the whole training procedure using early stopping, differing from our three-fold split with early stopping on the validation set, as described in Section~\ref{ssec:training}.
For both training regimes, both Gaze Integration variants achieve higher accuracies, with Cross-Attention performing better than the Two-Way approach.
For more details, see Table~\ref{tab:gaze_gnn}.

\begin{table}[h!]
    \centering
    \begin{tabular}{l|c|c}
    \hline
     Architecture & E.S. Test & E.S. Validation \\
     \hline
     \hline
     GazeGNN~\cite{wang_gazegnn_2024} & 83.18 & - \\
     GazeGNN~\cite{wang_gazegnn_2024} (repr.) & 82.94 & 71.02  \\
     Cross-Att.    & \textbf{87.96} & \textbf{84.11} \\
     Two-Way  & 85.78 & 82.80  \\
     \hline
\end{tabular}
    \caption{\textbf{GazeGNN comparison}: For better comparison, we report results on the CXR-Gaze test dataset using both evaluation strategies, early stopping on the test dataset as conducted by \citep{wang_gazegnn_2024} ("E.S. Test") and early stopping on a separate validation set ("E.S: Validation"). Under their training protocol, we are able to reproduce the results from Wang et al.~\citep{wang_gazegnn_2024}.}
    \label{tab:gaze_gnn}
\end{table}

\paragraph{SIIM-ACR Dataset}
For this dataset, both of our methods achieve accuracies close to the current state-of-art. 
Although Cross-Attention performs slightly worse than EG-ViT, the Two-Way Attention marginally improves classification accuracy to $86.40\%$.
In terms of AUC, our methods substantially outperform EG-ViT; however, the latter achieves a better F1-score.
In total, the two proposed FixationFormer variants match the performance of EG-ViT, though they do not surpass it.

\paragraph{Reflacx Data} 
The achieved accuracies on the Reflacx dataset indicate that  this task is the most challenging of the three.
In particular, F1 scores are substantially lower compared to both the CXR-Gaze and SIIM-ACR dataset.
This might be explained by the strong imbalance between classes, resulting in poor precision and recall for the under-represented categories, which leads to a low F1 score.
Cross-Attention outperforms Two-Way Attention, achieving an accuracy of $70.06\%$, while also showing a lower standard deviation across runs.
This might indicate that for this task, training the Two-Way Attention variant is more unstable.
Notably, both variants outperform GazeGNN.

In summary, the Cross-Attention variant seems to produce better and more consistent results compared to Two-Way Attention.
It performs better on the CXR-Gaze dataset than the Two-Way variant, achieves comparable accuracy on SIIM-ACR and is more stable during training on Reflacx while also achieving slightly better performance.

\subsection{Impact of Gaze} 
\begin{table}[h!]
    \centering
    \begin{tabular}{l|c|c|c}
    \hline
     Architecture & CXR-Gaze & SIIM-ACR & Reflacx \\
     \hline
     \hline
     Gaze Only          & 60.00 \scriptsize $\pm$ 2.67  & 79.20 \scriptsize $\pm$ 0.00  & 50.62 \scriptsize $\pm$ 0.67  \\
     Image Only         & 81.50 \scriptsize $\pm$ 1.54  & \textbf{86.32} \scriptsize $\pm$ 1.07  & 64.91 \scriptsize $\pm$ 1.15  \\
     \hline
     Cross-Att.    & \textbf{84.11} \scriptsize $\pm$ 1.14  & 84.96 \scriptsize $\pm$ 1.56  & \textbf{70.06} \scriptsize $\pm$ 0.93  \\
     Two-Way  & 82.80 \scriptsize $\pm$ 1.07  & \textbf{86.40} \scriptsize $\pm$ 1.56
     & 68.06 \scriptsize $\pm$ 2.32  \\
     \hline
\end{tabular}
    \caption{\textbf{Ablation Study}: Classification performance for models using only gaze or images, respectively, and a comparison against our proposed methods. 
    Mean accuracy and standard deviation over 5 runs are reported.}
    \label{tab:ablation}
\end{table}

\begin{figure*}[ht!]
\centering
      \includegraphics[width=\textwidth]{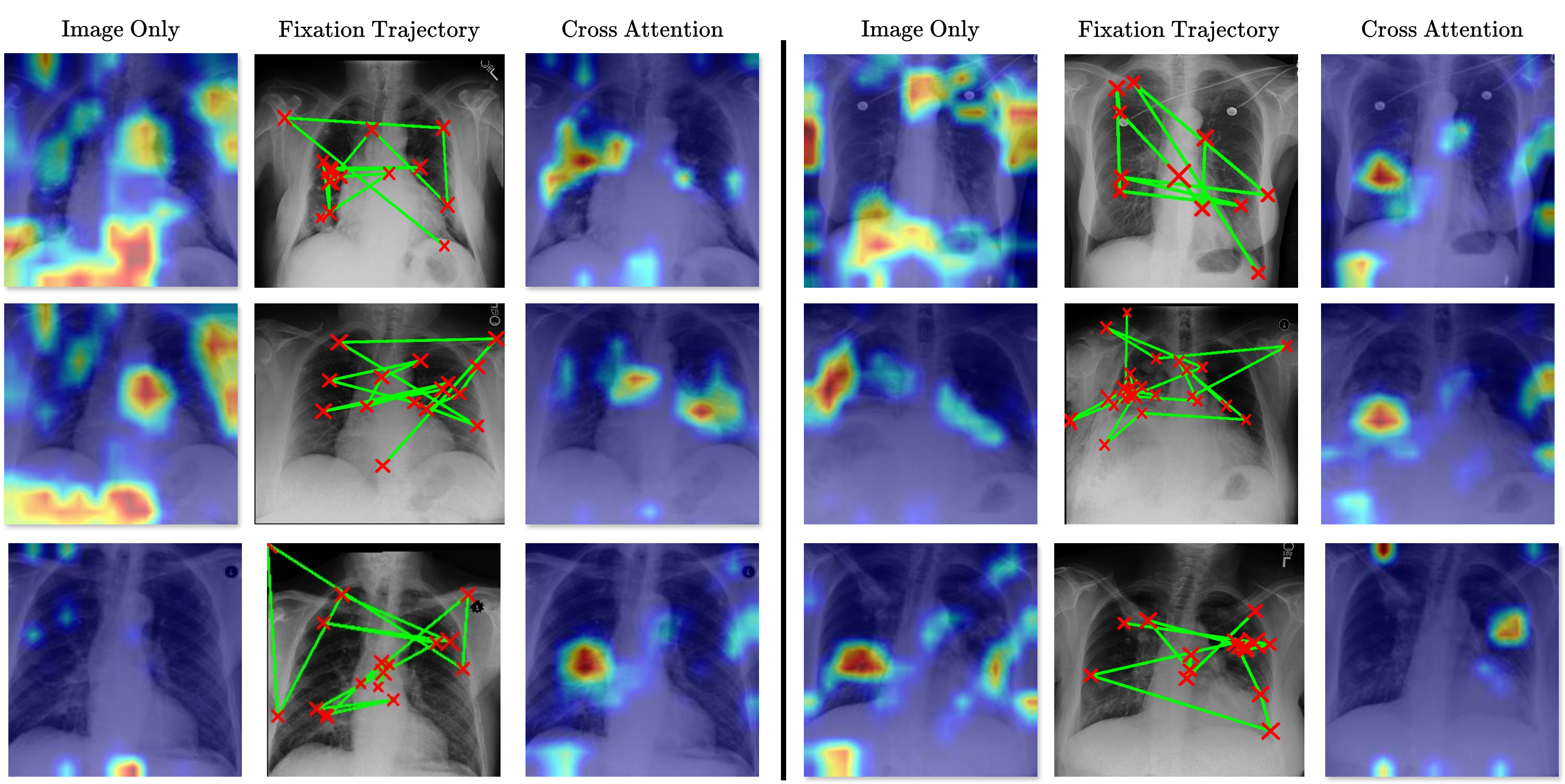}
\caption{\textbf{GradCAM-based attention visualization}: Comparison of model attention for correctly classified samples. 
Left: GradCAM image for the Image Only model. 
Center: Visualization of the expert fixation trajectory. 
Right: GradCAM image for the Cross-Attention model.} 
\label{fig:cam_comparison}
\end{figure*}

To study the impact of our gaze representation and the proposed integration methods, we conduct an ablation study.
The results are summarized in Table~\ref{tab:ablation}.

First, to assess the effectiveness of our proposed representation of gaze, we train a model using only the available gaze information and no images.
This model consists of a standard encoder-style Transformer with only self-attention and a linear head to predict class probabilities.
Results of this experiment are given in the first row of Table~\ref{tab:ablation}.
As expected, the performance of the gaze-only model significantly lags behind that of the other variants.
However, for both the CXR-Gaze and Reflacx dataset, we are able to achieve accuracies far above random guessing, indicating that the Transformer is able to capture meaningful semantics from our gaze token sequence.
Even though the Gaze Only accuracy on SIIM-ACR is high as well, the model was not able to capture relevant information from the gaze in this dataset.
All models trained on SIIM-ACR collapsed onto always predicting the overrepresented class, explaining the standard deviation of $0$.

In a second experiment, we compare our proposed architecture against a baseline using only the MGCA pretrained image encoder and no gaze.
The results for this model are depicted in the second row of Table~\ref{tab:ablation}.
For the CXR-Gaze dataset, adding gaze to the powerful MGCA image backbone yields modest accuracy gains: about $1\%$ for the Two-Way Attention variant and about $2.5\%$ for Cross-Attention.
In line with the collapse of the Gaze Only model on the SIIM-ACR dataset, adding gaze to the image encoder does not increase classification accuracy.
In contrast, for the challenging Reflacx dataset, adding gaze significantly improves performances using the Cross-Attention variant.
The Two-Way Attention mechanism also improves accuracy, but with higher variance.

\begin{table}[h!]
    \centering
    \begin{tabular}{l|c|c|c}
    \hline
     ImageNet Variant & CXR-Gaze & SIIM-ACR & Reflacx \\
     \hline
     \hline
     Image Only              & 64.49 & 79.20 & 59.68 \\
     Cross-Att.   & \textbf{72.90} & \textbf{82.80} & \textbf{61.28} \\
     Two-Way           & 63.55 & 80.00 & 58.08 \\
     \hline
\end{tabular}
    \caption{\textbf{ImageNet Backbone}: Classification accuracy reported for our method using an ImageNet backbone instead of the MGCA pretrained image encoder.}
    \label{tab:imagenet}
\end{table}

We also investigate the combination of our Gaze Integration Module with a weaker backbone. 
Specifically, we exchange the MGCA pretrained image encoder with a standard ImageNet ViT/B-32, and test the encoder standalone and with both variants of the Gaze Integration Module.
Results are summarized in Table~\ref{tab:imagenet}.
Interestingly, the two variants behave very differently when combined with this task-agnostic backbone.
While the Two-Way variant does not improve classification performance across all datasets, we see a clear improvement for the Cross-Attention gaze integration.
Notably, the performance on the CXR-Gaze dataset increases substantially when gaze is introduced for the ImageNet backbone, providing further evidence that the pre-trained MGCA backbone, when used with images only, is already highly effective for these datasets.


Finally, in Figure~\ref{fig:cam_comparison} we provide qualitative insights by visualizing attention maps for both the Image Only and Cross-Attention variant.
A comparison to the expert's gaze trajectory is also given.
We use GradCAM~\citep{Selvaraju_2019} to visualize the models attention on correctly classified samples.
Due to technical limitations, PyTorch currently does not support the extraction of attention weights when using Nested Tensors, which could be analysed with regards to the connection of image patches and fixation tokens as well.
For the model with gaze, attention maps reveal correlation with the gaze trajectory.
In particular, the gradient-based attention seems to be more focused and more consistently placed on anatomically relevant regions in comparison to the image-only variant.

\section{Conclusion}
\label{sec:conclusion}
In this work, we present FixationFormer, a transformer-based method for direct integration of gaze sequences into medical image analysis.
The key novelty lies in representing sequential fixation trajectories, directly derived from raw gaze data, as tokens that can be naturally integrated into a Transformer architecture.
These gaze tokens are fused with image features through two attention mechanisms, enabling different levels of interaction between gaze and image representations.

We evaluate our method on three chest X-ray classification tasks, achieving state-of-the-art performance on two datasets and matching the best results from previous methods on the third.
We further analyse the direct representation of a gaze trajectory as a token sequence.
Specifically, we show that with our gaze tokenisation, meaningful semantics are captured from the gaze data even without image context.
Additionally, we study the impact of gaze on classification performance.
We demonstrate that performance improves on two of the three examined datasets, irrespective of whether a strong task-specific image backbone or a standard ImageNet ViT is employed.
When employing the weaker backbone, the gap in performance between models with and without gaze integration widens substantially, providing evidence that FixationFormer can deliver performance improvements in situations with less favorable trade-offs between model and data complexity.

Interestingly, the Two-Way variant of our Gaze Integration module, which additionally allows gaze tokens to cross-attend to image features, performs worse and is more volatile than the one-way Cross-Attention approach. 
To further investigate this behavior, it would be interesting to directly analyze the attention weights learned within the cross-attention mechanism to compare both variants, as soon as this is supported for Nested Tensors.

As future work, the effectiveness of FixationFormer could be further examined through experiments on additional imaging modalities and tasks, which would represent a natural extension of this study. 
At present, suitable medical datasets remain rare. 
However, with the growing interest in gaze-based vision systems, particularly in the medical domain, it is reasonable to expect the release of additional datasets, which we plan to leverage to further evaluate the performance and potential extensions of FixationFormer.


\bibliographystyle{unsrtnat}


\begin{thebibliography}{36}
	\providecommand{\natexlab}[1]{#1}
	\providecommand{\url}[1]{\texttt{#1}}
	\expandafter\ifx\csname urlstyle\endcsname\relax
	\providecommand{\doi}[1]{doi: #1}\else
	\providecommand{\doi}{doi: \begingroup \urlstyle{rm}\Url}\fi
	
	\bibitem[Zhou et~al.(2018)Zhou, Siddiquee, Tajbakhsh, and
	Liang]{zhou2018unetnestedunetarchitecture}
	Zongwei Zhou, Md~Mahfuzur~Rahman Siddiquee, Nima Tajbakhsh, and Jianming Liang.
	\newblock Unet++: A nested u-net architecture for medical image segmentation,
	2018.
	\newblock URL \url{https://arxiv.org/abs/1807.10165}.
	
	\bibitem[Isensee et~al.(2021)Isensee, Jaeger, Kohl, Petersen, and
	Maier-Hein]{isensee2021nnunet}
	Fabian Isensee, Paul~F. Jaeger, Simon A.~A. Kohl, Jens Petersen, and Klaus~H.
	Maier-Hein.
	\newblock nnu-net: a self-configuring method for deep learning-based biomedical
	image segmentation.
	\newblock \emph{Nature Methods}, 2021.
	\newblock \doi{10.1038/s41592-020-01008-z}.
	\newblock URL \url{https://doi.org/10.1038/s41592-020-01008-z}.
	
	\bibitem[Karargyris et~al.(2021)Karargyris, Kashyap, Lourentzou, Wu, Sharma,
	Tong, Abedin, Beymer, Mukherjee, Krupinski, and
	Moradi]{karargyris_creation_2021}
	Alexandros Karargyris, Satyananda Kashyap, Ismini Lourentzou, Joy~T. Wu, Arjun
	Sharma, Matthew Tong, Shafiq Abedin, David Beymer, Vandana Mukherjee,
	Elizabeth~A. Krupinski, and Mehdi Moradi.
	\newblock Creation and validation of a chest {X}-ray dataset with eye-tracking
	and report dictation for {AI} development.
	\newblock \emph{Scientific Data}, 8, 2021.
	\newblock ISSN 2052-4463.
	\newblock \doi{10.1038/s41597-021-00863-5}.
	\newblock URL \url{https://www.nature.com/articles/s41597-021-00863-5}.
	
	\bibitem[Zhu et~al.(2022)Zhu, Salcudean, and Rohling]{zhu_gaze-guided_2022}
	Hongzhi Zhu, Septimiu Salcudean, and Robert Rohling.
	\newblock Gaze-{Guided} {Class} {Activation} {Mapping}: {Leveraging} {Human}
	{Attention} for {Network} {Attention} in {Chest} {X}-rays {Classification},
	2022.
	\newblock URL \url{http://arxiv.org/abs/2202.07107}.
	
	\bibitem[Saab et~al.(2021)Saab, Hooper, Sohoni, Parmar, Pogatchnik, Wu,
	Dunnmon, Zhang, Rubin, and Ré]{saab_observational_2021}
	Khaled Saab, Sarah~M. Hooper, Nimit~S. Sohoni, Jupinder Parmar, Brian
	Pogatchnik, Sen Wu, Jared~A. Dunnmon, Hongyang~R. Zhang, Daniel Rubin, and
	Christopher Ré.
	\newblock Observational {Supervision} for {Medical} {Image} {Classification}
	{Using} {Gaze} {Data}.
	\newblock In \emph{Medical {Image} {Computing} and {Computer} {Assisted}
		{Intervention} – {MICCAI} 2021}, Cham, 2021. Springer International
	Publishing.
	\newblock ISBN 978-3-030-87196-3.
	\newblock \doi{10.1007/978-3-030-87196-3_56}.
	
	\bibitem[Wang et~al.(2022{\natexlab{a}})Wang, Ouyang, Liu, Wang, and
	Shen]{wang_follow_2022}
	Sheng Wang, Xi~Ouyang, Tianming Liu, Qian Wang, and Dinggang Shen.
	\newblock Follow {My} {Eye}: {Using} {Gaze} to {Supervise} {Computer}-{Aided}
	{Diagnosis}.
	\newblock \emph{IEEE Transactions on Medical Imaging}, 41, 2022{\natexlab{a}}.
	\newblock ISSN 0278-0062, 1558-254X.
	\newblock \doi{10.1109/TMI.2022.3146973}.
	\newblock URL \url{http://arxiv.org/abs/2204.02976}.
	
	\bibitem[Ma et~al.(2022)Ma, Zhao, Chen, Zhang, Xiao, Dai, Liu, Wu, Liu, Wang,
	Gao, Li, Jiang, Zhang, Wang, Shen, Zhu, and Liu]{ma_eye-gaze-guided_2022}
	Chong Ma, Lin Zhao, Yuzhong Chen, Lu~Zhang, Zhenxiang Xiao, Haixing Dai, David
	Liu, Zihao Wu, Zhengliang Liu, Sheng Wang, Jiaxing Gao, Changhe Li, Xi~Jiang,
	Tuo Zhang, Qian Wang, Dinggang Shen, Dajiang Zhu, and Tianming Liu.
	\newblock Eye-gaze-guided {Vision} {Transformer} for {Rectifying} {Shortcut}
	{Learning}, 2022.
	\newblock URL \url{http://arxiv.org/abs/2205.12466}.
	
	\bibitem[Wang et~al.(2024)Wang, Pan, Aboah, Zhang, Keles, Torigian, Turkbey,
	Krupinski, Udupa, and Bagci]{wang_gazegnn_2024}
	Bin Wang, Hongyi Pan, Armstrong Aboah, Zheyuan Zhang, Elif Keles, Drew
	Torigian, Baris Turkbey, Elizabeth Krupinski, Jayaram Udupa, and Ulas Bagci.
	\newblock {GazeGNN}: {A} {Gaze}-{Guided} {Graph} {Neural} {Network} for {Chest}
	{X}-ray {Classification}.
	\newblock 2024.
	\newblock ISBN 979-8-3503-1892-0.
	\newblock \doi{10.1109/WACV57701.2024.00219}.
	\newblock URL \url{https://ieeexplore.ieee.org/document/10484310/}.
	
	\bibitem[Dosovitskiy et~al.(2021)Dosovitskiy, Beyer, Kolesnikov, Weissenborn,
	Zhai, Unterthiner, Dehghani, Minderer, Heigold, Gelly, Uszkoreit, and
	Houlsby]{dosovitskiy2021imageworth16x16words}
	Alexey Dosovitskiy, Lucas Beyer, Alexander Kolesnikov, Dirk Weissenborn,
	Xiaohua Zhai, Thomas Unterthiner, Mostafa Dehghani, Matthias Minderer, Georg
	Heigold, Sylvain Gelly, Jakob Uszkoreit, and Neil Houlsby.
	\newblock An image is worth 16x16 words: Transformers for image recognition at
	scale, 2021.
	\newblock URL \url{https://arxiv.org/abs/2010.11929}.
	
	\bibitem[Wang et~al.(2023)Wang, Aboah, Zhang, and Bagci]{wang2023gazesam}
	Bin Wang, Armstrong Aboah, Zheyuan Zhang, and Ulas Bagci.
	\newblock Gazesam: What you see is what you segment, 2023.
	
	\bibitem[Beckmann et~al.(2023)Beckmann, Kockwelp, Gromoll, Kiefer, and
	Risse]{beckmann2023sam}
	Daniel Beckmann, Jacqueline Kockwelp, J{\"o}rg Gromoll, Friedemann Kiefer, and
	Benjamin Risse.
	\newblock Sam meets gaze: Passive eye tracking for prompt-based instance
	segmentation.
	\newblock In \emph{NeuRIPS 2023 Workshop on Gaze Meets ML}, 2023.
	
	\bibitem[Khaertdinova et~al.(2024)Khaertdinova, Pershin, Shmykova, and
	Ibragimov]{khaertdinova_gaze-assisted_2024}
	Leila Khaertdinova, Ilya Pershin, Tatiana Shmykova, and Bulat Ibragimov.
	\newblock Gaze-{Assisted} {Medical} {Image} {Segmentation}, 2024.
	\newblock URL \url{http://arxiv.org/abs/2410.17920}.
	
	\bibitem[Shmykova et~al.(2025)Shmykova, Khaertdinova, and
	Pershin]{shmykova_zero-shot_2025}
	Tatyana Shmykova, Leila Khaertdinova, and Ilya Pershin.
	\newblock Zero-{Shot} {Gaze}-based {Volumetric} {Medical} {Image}
	{Segmentation}, 2025.
	\newblock URL \url{https://arxiv.org/abs/2505.15256v1}.
	
	\bibitem[Johnson et~al.(2019)Johnson, Pollard, Berkowitz, Greenbaum, Lungren,
	Deng, Mark, and Horng]{johnson_mimic-cxr_2019}
	Alistair E.~W. Johnson, Tom~J. Pollard, Seth~J. Berkowitz, Nathaniel~R.
	Greenbaum, Matthew~P. Lungren, Chih-ying Deng, Roger~G. Mark, and Steven
	Horng.
	\newblock {MIMIC}-{CXR}, a de-identified publicly available database of chest
	radiographs with free-text reports.
	\newblock \emph{Scientific Data}, 6, 2019.
	\newblock ISSN 2052-4463.
	\newblock \doi{10.1038/s41597-019-0322-0}.
	\newblock URL \url{https://doi.org/10.1038/s41597-019-0322-0}.
	
	\bibitem[Koorathota et~al.(2025)Koorathota, Papadopoulos, Ma, Kumar, Sun,
	Mittal, Adelman, and Sajda]{koorathota_gaze-informed_2025}
	Sharath Koorathota, Nikolas Papadopoulos, Jia~Li Ma, Shruti Kumar, Xiaoxiao
	Sun, Arunesh Mittal, Patrick Adelman, and Paul Sajda.
	\newblock Gaze-{Informed} {Vision} {Transformers}: {Predicting} {Driving}
	{Decisions} {Under} {Uncertainty}, 2025.
	\newblock URL \url{http://arxiv.org/abs/2308.13969}.
	
	\bibitem[Hu et~al.(2025)Hu, Tong, Gao, Zeng, Yan, and Li]{hu_gazevit_2025}
	Yidong Hu, Li~Tong, Yuanlong Gao, Ying Zeng, Bin Yan, and Zhongrui Li.
	\newblock {GazeViT}: {A} gaze-guided hybrid attention vision transformer for
	cross-view matching of street-to-aerial images.
	\newblock 191, 2025.
	\newblock ISSN 0167-8655.
	\newblock \doi{10.1016/j.patrec.2025.03.012}.
	\newblock URL
	\url{https://www.sciencedirect.com/science/article/pii/S0167865525000935}.
	
	\bibitem[Zhou et~al.(2015)Zhou, Khosla, Lapedriza, Oliva, and
	Torralba]{zhou2015learningdeepfeaturesdiscriminative}
	Bolei Zhou, Aditya Khosla, Agata Lapedriza, Aude Oliva, and Antonio Torralba.
	\newblock Learning deep features for discriminative localization, 2015.
	\newblock URL \url{https://arxiv.org/abs/1512.04150}.
	
	\bibitem[Ma et~al.(2024{\natexlab{a}})Ma, Jiang, Chen, Li, Wu, Yu, Liu, Guo,
	Zhu, Zhang, Shen, Liu, and Li]{ma_eye-gaze_2024}
	Chong Ma, Hanqi Jiang, Wenting Chen, Yiwei Li, Zihao Wu, Xiaowei Yu, Zhengliang
	Liu, Lei Guo, Dajiang Zhu, Tuo Zhang, Dinggang Shen, Tianming Liu, and Xiang
	Li.
	\newblock Eye-gaze {Guided} {Multi}-modal {Alignment} for {Medical}
	{Representation} {Learning}, 2024{\natexlab{a}}.
	\newblock URL \url{http://arxiv.org/abs/2403.12416}.
	
	\bibitem[Radford et~al.(2021)Radford, Kim, Hallacy, Ramesh, Goh, Agarwal,
	Sastry, Askell, Mishkin, Clark, Krueger, and
	Sutskever]{radford2021learningtransferablevisualmodels}
	Alec Radford, Jong~Wook Kim, Chris Hallacy, Aditya Ramesh, Gabriel Goh,
	Sandhini Agarwal, Girish Sastry, Amanda Askell, Pamela Mishkin, Jack Clark,
	Gretchen Krueger, and Ilya Sutskever.
	\newblock Learning transferable visual models from natural language
	supervision, 2021.
	\newblock URL \url{https://arxiv.org/abs/2103.00020}.
	
	\bibitem[Wang et~al.(2025)Wang, Zhao, Shen, Wang, Wang, and
	Shen]{wang_improving_2025}
	Sheng Wang, Zihao Zhao, Zhenrong Shen, Bin Wang, Qian Wang, and Dinggang Shen.
	\newblock Improving {Self}-{Supervised} {Medical} {Image} {Pre}-{Training} by
	{Early} {Alignment} with {Human} {Eye} {Gaze} {Information}.
	\newblock 2025.
	\newblock \doi{10.1109/TMI.2025.3528965}.
	\newblock URL \url{https://ieeexplore.ieee.org/abstract/document/10839445}.
	
	\bibitem[Kirillov et~al.(2023)Kirillov, Mintun, Ravi, Mao, Rolland, Gustafson,
	Xiao, Whitehead, Berg, Lo, Dollár, and Girshick]{kirillov2023segment}
	Alexander Kirillov, Eric Mintun, Nikhila Ravi, Hanzi Mao, Chloe Rolland, Laura
	Gustafson, Tete Xiao, Spencer Whitehead, Alexander~C. Berg, Wan-Yen Lo, Piotr
	Dollár, and Ross Girshick.
	\newblock Segment anything, 2023.
	\newblock URL \url{https://arxiv.org/abs/2304.02643}.
	
	\bibitem[Ma et~al.(2024{\natexlab{b}})Ma, He, Li, Han, You, and Wang]{MedSAM}
	Jun Ma, Yuting He, Feifei Li, Lin Han, Chenyu You, and Bo~Wang.
	\newblock Segment anything in medical images.
	\newblock \emph{Nature Communications}, 15:\penalty0 654, 2024{\natexlab{b}}.
	
	\bibitem[Ma et~al.(2025)Ma, Yang, Kim, Chen, Baharoon, Fallahpour, Asakereh,
	Lyu, and Wang]{MedSAM2}
	Jun Ma, Zongxin Yang, Sumin Kim, Bihui Chen, Mohammed Baharoon, Adibvafa
	Fallahpour, Reza Asakereh, Hongwei Lyu, and Bo~Wang.
	\newblock Medsam2: Segment anything in 3d medical images and videos.
	\newblock \emph{arXiv preprint arXiv:2504.03600}, 2025.
	
	\bibitem[Pham et~al.(2025)Pham, Brecheisen, Wu, Nguyen, Deng, Adjeroh, Doretto,
	Choudhary, and Le]{pham2025itpctrl}
	Trong-Thang Pham, Jacob Brecheisen, Carol~C Wu, Hien Nguyen, Zhigang Deng,
	Donald Adjeroh, Gianfranco Doretto, Arabinda Choudhary, and Ngan Le.
	\newblock Itpctrl-ai: End-to-end interpretable and controllable artificial
	intelligence by modeling radiologists’ intentions.
	\newblock \emph{Artificial Intelligence in Medicine}, 160:\penalty0 103054,
	2025.
	
	\bibitem[Chen et~al.(2026)Chen, Liu, and Song]{chen2026gaze}
	Zihui Chen, Zhi Liu, and Yingjie Song.
	\newblock Gaze-guided vision transformer for chest x-ray image classification.
	\newblock \emph{Biomedical Signal Processing and Control}, 111:\penalty0
	108298, 2026.
	
	\bibitem[Bhattacharya et~al.(2022)Bhattacharya, Jain, and
	Prasanna]{bhattacharya2022radiotransformer}
	Moinak Bhattacharya, Shubham Jain, and Prateek Prasanna.
	\newblock Radiotransformer: a cascaded global-focal transformer for visual
	attention--guided disease classification.
	\newblock In \emph{European Conference on Computer Vision}, pages 679--698.
	Springer, 2022.
	
	\bibitem[Zhu et~al.(2023)Zhu, Chen, and
	Yang]{zhu2023understandingvittrainsbadly}
	Haoran Zhu, Boyuan Chen, and Carter Yang.
	\newblock Understanding why vit trains badly on small datasets: An intuitive
	perspective, 2023.
	\newblock URL \url{https://arxiv.org/abs/2302.03751}.
	
	\bibitem[Lee et~al.(2021)Lee, Lee, and
	Song]{lee2021visiontransformersmallsizedatasets}
	Seung~Hoon Lee, Seunghyun Lee, and Byung~Cheol Song.
	\newblock Vision transformer for small-size datasets, 2021.
	\newblock URL \url{https://arxiv.org/abs/2112.13492}.
	
	\bibitem[Wang et~al.(2022{\natexlab{b}})Wang, Zhou, Wang, Vardhanabhuti, and
	Yu]{wang_multi-granularity_2022}
	Fuying Wang, Yuyin Zhou, Shujun Wang, Varut Vardhanabhuti, and Lequan Yu.
	\newblock Multi-{Granularity} {Cross}-modal {Alignment} for {Generalized}
	{Medical} {Visual} {Representation} {Learning}, 2022{\natexlab{b}}.
	\newblock URL \url{http://arxiv.org/abs/2210.06044}.
	
	\bibitem[Vaswani et~al.(20217)Vaswani, Shazeer, Parmar, Uszkoreit, Jones,
	Gomez, Kaiser, and Polosukhin]{vaswani2017attentionneed}
	Ashish Vaswani, Noam Shazeer, Niki Parmar, Jakob Uszkoreit, Llion Jones,
	Aidan~N. Gomez, Lukasz Kaiser, and Illia Polosukhin.
	\newblock Attention is all you need, 20217.
	\newblock URL \url{https://arxiv.org/abs/1706.03762}.
	
	\bibitem[Zawacki et~al.(2019)Zawacki, Wu, Shih, Elliott, Fomitchev, Hussain,
	ParasLakhani, Culliton, and Bao]{siim-acr-pneumothorax-segmentation}
	Anna Zawacki, Carol Wu, George Shih, Julia Elliott, Mikhail Fomitchev, Mohannad
	Hussain, ParasLakhani, Phil Culliton, and Shunxing Bao.
	\newblock Siim-acr pneumothorax segmentation.
	\newblock
	\url{https://kaggle.com/competitions/siim-acr-pneumothorax-segmentation},
	2019.
	\newblock Kaggle.
	
	\bibitem[Bigolin~Lanfredi et~al.(2022)Bigolin~Lanfredi, Zhang, Auffermann,
	Chan, Duong, Srikumar, Drew, Schroeder, and
	Tasdizen]{bigolin_lanfredi_reflacx_2022}
	Ricardo Bigolin~Lanfredi, Mingyuan Zhang, William~F. Auffermann, Jessica Chan,
	Phuong-Anh~T. Duong, Vivek Srikumar, Trafton Drew, Joyce~D. Schroeder, and
	Tolga Tasdizen.
	\newblock {REFLACX}, a dataset of reports and eye-tracking data for
	localization of abnormalities in chest x-rays.
	\newblock \emph{Scientific Data}, 9, 2022.
	\newblock ISSN 2052-4463.
	\newblock \doi{10.1038/s41597-022-01441-z}.
	\newblock URL \url{https://www.nature.com/articles/s41597-022-01441-z}.
	
	\bibitem[Loshchilov and
	Hutter(2019)]{loshchilov2019decoupledweightdecayregularization}
	Ilya Loshchilov and Frank Hutter.
	\newblock Decoupled weight decay regularization, 2019.
	\newblock URL \url{https://arxiv.org/abs/1711.05101}.
	
	\bibitem[Hu et~al.(2021)Hu, Shen, Wallis, Allen-Zhu, Li, Wang, Wang, and
	Chen]{hu2021loralowrankadaptationlarge}
	Edward~J. Hu, Yelong Shen, Phillip Wallis, Zeyuan Allen-Zhu, Yuanzhi Li, Shean
	Wang, Lu~Wang, and Weizhu Chen.
	\newblock Lora: Low-rank adaptation of large language models, 2021.
	\newblock URL \url{https://arxiv.org/abs/2106.09685}.
	
	\bibitem[Paszke et~al.(2019)Paszke, Gross, Massa, Lerer, Bradbury, Chanan,
	Killeen, Lin, Gimelshein, Antiga, Desmaison, Köpf, Yang, DeVito, Raison,
	Tejani, Chilamkurthy, Steiner, Fang, Bai, and
	Chintala]{paszke2019pytorchimperativestylehighperformance}
	Adam Paszke, Sam Gross, Francisco Massa, Adam Lerer, James Bradbury, Gregory
	Chanan, Trevor Killeen, Zeming Lin, Natalia Gimelshein, Luca Antiga, Alban
	Desmaison, Andreas Köpf, Edward Yang, Zach DeVito, Martin Raison, Alykhan
	Tejani, Sasank Chilamkurthy, Benoit Steiner, Lu~Fang, Junjie Bai, and Soumith
	Chintala.
	\newblock Pytorch: An imperative style, high-performance deep learning library,
	2019.
	\newblock URL \url{https://arxiv.org/abs/1912.01703}.
	
	\bibitem[Selvaraju et~al.(2019)Selvaraju, Cogswell, Das, Vedantam, Parikh, and
	Batra]{Selvaraju_2019}
	Ramprasaath~R. Selvaraju, Michael Cogswell, Abhishek Das, Ramakrishna Vedantam,
	Devi Parikh, and Dhruv Batra.
	\newblock Grad-cam: Visual explanations from deep networks via gradient-based
	localization.
	\newblock \emph{International Journal of Computer Vision}, 2019.
	\newblock \doi{10.1007/s11263-019-01228-7}.
	\newblock URL \url{http://dx.doi.org/10.1007/s11263-019-01228-7}.
	
\end{thebibliography}

\end{document}